\documentclass[runningheads]{llncs}
\usepackage[T1]{fontenc}
\usepackage{cite}
\usepackage{amsmath,amssymb,amsfonts}
\usepackage{algorithm}
\usepackage{algorithmicx}
\usepackage[noend]{algpseudocode}
\usepackage[caption=false]{subfig}

\usepackage{graphicx}
\usepackage{textcomp}
\usepackage{xcolor}
\usepackage{tcolorbox}
\usepackage{graphicx}

\usepackage{hyperref}

\usepackage{comment}
\usepackage{todonotes}

\usepackage{latexsym}
\usepackage{subfig}
\usepackage{float}

\usepackage{colortbl}
\usepackage{multirow}
\usepackage{adjustbox}

\usepackage{chngcntr}
\usepackage{booktabs}

\definecolor{lightgray}{gray}{0.9}
\definecolor{softblue}{RGB}{70, 130, 180} 
\definecolor{softgreen}{RGB}{34, 139, 34} 
\definecolor{softred}{RGB}{220, 20, 60} 

\graphicspath{{figs/}}

\newtheorem{heuristic}{Heuristic Group}

\begin{document}
\title{Optimization of Activity Batching Policies in Business Processes}

%
\titlerunning{Optimization of Batching Policies in Business Processes}
%
\author{Orlenys L\'opez-Pintado\inst{1} \and
Jannis Rosenbaum\inst{1} \and
Marlon Dumas\inst{1}}

\authorrunning{O. L\'opez-Pintado, J. Rosenbaum and M. Dumas}
\institute{University of Tartu, Estonia \\
\email{\{orlenys.lopez.pintado, jannis.rosenbaum, marlon.dumas\}@ut.ee}}

\maketitle              
\begin{abstract}
In business processes, activity batching refers to packing multiple activity instances for joint execution. Batching allows managers to trade off cost and processing effort against waiting time. Larger and less frequent batches may lower costs by reducing processing effort and amortizing fixed costs, but they create longer waiting times. In contrast, smaller and more frequent batches reduce waiting times but increase fixed costs and processing effort. A batching policy defines how activity instances are grouped into batches and when each batch is activated. This paper addresses the problem of discovering batching policies that strike optimal trade-offs between waiting time, processing effort, and cost. The paper proposes a Pareto optimization approach that starts from a given set (possibly empty) of activity batching policies and generates alternative policies for each batched activity via intervention heuristics. Each heuristic identifies an opportunity to improve an activity's batching policy with respect to a metric (waiting time, processing time, cost, or resource utilization) and an associated adjustment to the activity's batching policy (the intervention). The impact of each intervention is evaluated via simulation. The intervention heuristics are embedded in an optimization meta-heuristic that triggers interventions to iteratively update the Pareto front of the interventions identified so far. The paper considers three meta-heuristics: hill-climbing, simulated annealing, and reinforcement learning. An experimental evaluation compares the proposed approach based on intervention heuristics against the same (non-heuristic guided) meta-heuristics baseline regarding convergence, diversity, and cycle time gain of Pareto-optimal policies.


\end{abstract}

\section{Introduction}
\label{sec:introduction}


In a business process, batching involves grouping multiple instances of the same activity for joint execution~\cite{PufahlW19}. Batching may reduce processing effort (e.g., by reducing context-switching) and costs (e.g., by amortizing fixed fees, such as transportation fees). Conversely, batching leads to higher waiting times, as activity instances need to wait until their corresponding batch is activated. For example, in a medical laboratory, testing each blood sample as soon as it arrives may lead to lower waiting times, assuming a test is completed before the following sample arrives. However, it is costly because the machine's operating expenses depend on the number of runs. Meanwhile, processing samples in batches lowers the processing time and the cost per test, but it makes patients wait longer. Similar trade-offs occur in logistics and manufacturing processes, among others.




Existing approaches to batch optimization in business processes~\cite{PufahlBW14} focus on optimizing batch sizes. In practice, though, batching policies may (also) involve time-based activation rules, e.g., a batch is activated ``every day at 8:00 AM'' or ``at most 12 hours after batch creation'' (time-to-live timeout) or ``after two hours if no new instance arrives'' (inactivity timeout). 




In this setting, this paper presents an automated approach to optimize batching policies that involve batch activation rules combining batch size and time-dependent conditions. Recognizing that batching policies introduce a trade-off between waiting time, processing effort, and fixed activity cost, the proposal adopts a Pareto optimization approach: Rather than discovering one single solution, it discovers a set of Pareto-optimal solutions, thus providing a spectrum of options to the user. Furthermore, the approach recognizes that batching policies may be attached to multiple (potentially all) activities in a process, so adjusting the batching policy for one activity may affect the effectiveness of batching policies attached to other activities. Accordingly, the approach optimizes the activity batching policies holistically, seeking solutions that optimize waiting time, processing effort, and cost at the process level.

The proposal comprises 19 intervention heuristics that can be used to generate new (potentially better) batch activation rules or update existing ones. Each heuristic identifies a situation where an adjustment may be applied to a batch activation rule to improve waiting time, processing time, fixed cost-per-instance, or resource utilization. The quality of each newly generated batch activation rule is evaluated regarding its impact on waiting time, processing time, and cost and added to the Pareto front if it dominates previously found solutions. 

The proposed heuristics can be used as perturbation heuristics within a meta-heuristic. The paper considers three meta-heuristics: Hill-Climbing (local optimization), Simulated Annealing (global optimization stochastically attempting non-optimal solutions), and Reinforcement Learning (global adaptative search with trial-and-error learning).  
Experiments show that, for each of these meta-heuristics, the proposed heuristic interventions lead to better Pareto fronts than the same meta-heuristic coupled with a standard random perturbation function over neighbor solutions, w.r.t., well-accepted quality metrics for Pareto fronts (convergence, diversity, cycle-time gain per process case).  


\section{Related Work}
\label{sec:related}

Liu \& Hu~\cite{LiuH07} propose grouping activities into Batch Processing Areas (BPAs), where instances accumulate until reaching a size threshold. Although the authors suggest optimizing this threshold via simulation, no method is provided. Pufahl et al.~\cite{PufahlMW2014} adopt a similar model based on \emph{batch regions}. In their model, the batch activation conditions depend both on batch size and time since batch creation (time-to-live timeout). Similarly, Natschl\"ager et al.~\cite{NatschlagerBGB15} consider batching policies based on batch size, time-to-live timeouts, and inactivity timeouts. Pufahl \& Karastoyanova~\cite{PufahlK18} propose a flexible batch execution model that, in addition to supporting batch activation rules, allows batches to be triggered based on contextual conditions or external triggers. However, none of the above studies deals with optimizing the batch activation rules.


Pflug \& Rinderle-Ma~\cite{PflugR16} propose a model wherein activity instances are classified at runtime based on their data attributes. Activity instances in the same class are processed in a batch. A batch is processed whenever a resource becomes available. This approach does not seek to optimize the batching policy.

Pufahl et al.~\cite{PufahlBW14} outline an approach to optimize the batch size threshold w.r.t.\ a cost function that combines the ``cost of waiting'' plus the processing effort. The cost of using a given batch size threshold is estimated via queuing theory. The approach uses a grid (exhaustive) search, which is only practical for small search spaces. This approach does not consider batch activation rules that depend on time bounds (only batch size), and it optimizes each batch activity separately. In the presence of multiple batch activities in a process, the batch activation rule of an activity affects the time when cases reach other downstream activities. Hence, optimizing the activation rules of each activity separately may lead to suboptimal results. Accordingly, our work optimizes the batch activation rules of all batched activities in a process simultaneously.

Pufahl \& Weske~\cite{PufahlW19} propose a batch processing model unifying prior approaches. In this model, batch activation rules include three types of conditions: (1) batch size, (2) time-to-live or inactivity timeout, and (3) circadian time, e.g., a batch activates on Monday at 8:00 AM following its creation time. We adopt/extend this latter approach and propose a method to find optimal activation rules w.r.t., waiting time, processing effort, and cost. 


Other studies consider the related problem of optimizing batch sizes and batch-to-resource assignment in production systems consisting of multiple workstations~\cite {RabtaR12,castilloG15,AkhtarHS19}. To explore the search space, these studies use queuing theory to estimate the waiting times generated by different batch sizes and meta-heuristics (e.g., genetic algorithms). These studies, however, do not consider batch activation policies that also depend on timeouts and circadian activation time.







\section{Batching Model \& Intervention Heuristics}
\label{sct:heuristic-model}

\subsection{Batching Model}
\label{sct:batch_model}

Batching is a scheduling mechanism that groups and executes multiple instances of the same activity together in a business process. Definitions~\ref{def:notation}-~\ref{def:batch_activation} formalize the batching model (a.k.a., batching policy) used in this paper, adapting the concepts from~\cite{MartinSMDC21, LashkevichMCD22}. Specifically, we incorporate the batching cost concept and formalize the batching activation rules supported by our approach.

\begin{definition}[General Notations]\label{def:notation}  
Let $\mathcal{A}$ be the set of activities in a business process, and let $\mathcal{I}_a$ denote the set of instances of each activity $a \in \mathcal{A}$, where each instance $i \in \mathcal{I}_a$ represents an execution of activity $a$. We define the following time-related functions for any instance \( i \): $\tau_e(i)$ represents the \textbf{enablement time}, when the activity instance becomes available for execution, $\tau_s(i)$ the \textbf{start time}, when execution begins, and $\tau_c(i)$ the \textbf{completion time} when it finishes. Let $\mathcal{R}$ be the set of resources, human or not, who can execute activity instances.
\end{definition}

\begin{definition}[Activity Batching]\label{def:batch} A \textbf{batch} is a tuple $\mathcal{B} = (\mathcal{I}^\mathcal{B}_a, r, \tau_s^\mathcal{B}, \tau_c^\mathcal{B}, \beta, \kappa)$, where $\mathcal{I}^B_a = \{i_1, i_2, \dots, i_n\} \subseteq \mathcal{I}_a$ is the set of batched instances of activity $a$; $r \in \mathcal{R}$ is the resource executing the batch; $\tau_s^\mathcal{B}$ and $\tau_c^\mathcal{B}$ are the start and completion times of the batch, and $\beta$ is the batching type. If $\beta$ is \textbf{sequential}, instances execute one after another, i.e., $\tau_s(i_l) = \tau_c(i_{l-1})$, for all $i_l \in \mathcal{I}^B_a$ with $l > 1$. If $\beta$ is \textbf{parallel}, all instances start and end together, i.e., $\tau_s(i_j) = \tau_s(i_l)$ and $\tau_c(i_j) = \tau_c(i_l)$ for all $i_j, i_l \in \mathcal{I}^B_a$. Each $i_l \in \mathcal{I}^B_a$ may belong to a different process case but follows the same batch constraints. The batch execution cost is $\kappa = \kappa_f + \kappa_v(\lvert \mathcal{I}_a^B \rvert) + \kappa_r(r, \tau_c^\mathcal{B} - \tau_s^\mathcal{B})$, where $\kappa_f$ is a fixed execution cost, $\kappa_v$ depends on batch size, and $\kappa_r$ depends on resource $r$ and execution time $\tau_c^\mathcal{B} - \tau_s^\mathcal{B}$. We use the notation $\mathcal{B}^{\mathcal{P}}_a$ to denote the set of all batches of a given activity $a$ executed within a business process execution $\mathcal{P}$. 
\end{definition}

\begin{definition}\label{def:batch_activation}
A \textbf{batch activation rule} is a boolean function associated to an activity $a \in \mathcal{A}$ that determines when a batch $\mathcal{B} = (\mathcal{I}^\mathcal{B}_a, r, \tau_s^\mathcal{B}, \tau_c^\mathcal{B}, \beta, \kappa)$ is started, defined as follows:
$$
\alpha(\sigma_t) = \bigvee_{j=1}^{m} \Big( \bigwedge_{x \in G_j} c^x(\sigma_t) \Big),
$$
where \( \sigma_t \) is the \textbf{process state} at time \( t \) and each \( c^x(\sigma_t) \) is an activation condition. Each conjunction (\(\land\)) requires that all conditions in a group \( G_j \) hold simultaneously, while the disjunction (\(\lor\)) allows any group of activation conditions \( G_j \) to trigger the batch. The activation conditions include:
\begin{itemize}
    \item \textbf{Size-based activation}, \( c^{\text{size}}(\sigma_t) \): The batch starts if the number of enabled activity instances waiting reaches a threshold, i.e., $ c^{\text{size}}(\sigma_t) \equiv \lvert \mathcal{I}_a^\mathcal{B} \rvert \geq \theta_v$.
    
    \item \textbf{(Waiting) Time-based activation}:
    \begin{itemize}
        \item \textbf{Since first (time-to-live)}, \( c^{\text{wt-first}}(\sigma_t) \): The earliest activity instance waiting has waited at least \(\theta_f\) time units, $ c^{\text{wt-first}}(\sigma_t) \equiv t - \tau_e(i_1) \geq \theta_f$.
        \item \textbf{Since last (inactivity)}, \( c^{\text{wt-last}}(\sigma_t) \): The latest activity instance waiting has waited at least \(\theta_l\) time units, i.e., $c^{\text{wt-last}}(\sigma_t) \equiv t - \tau_e(i_n) \geq \theta_l$.
    \end{itemize}
    
    \item \textbf{Scheduled-based activation}:
    \begin{itemize}
        \item \textbf{Daily hour rule}, \( c^{\text{hour}}(\sigma_t) \): The batch starts if the hour extracted from \( \sigma_t \) belongs to a predefined set of activation hours \( T_{\text{allowed}} \subseteq \{0,1,\dots,23\} \), $c^{\text{hour}}(\sigma_t) \equiv \text{hour}(\sigma_t) \in T_{\text{allowed}}$.

        \item \textbf{Day of the week rule}, \( c^{\text{day}}(\sigma_t) \): The batch starts if the week-day extracted from \( \sigma_t \) belongs to a predefined set of allowed days \( D_{\text{allowed}} \subseteq \{\text{Monday}, \dots, \text{Sunday} \} \), $c^{\text{day}}(\sigma_t) \equiv \text{weekday}(\sigma_t) \in D_{\text{allowed}}$.
    \end{itemize}
\end{itemize}
Enabled activity instances accumulate in $\mathcal{I}_a^\mathcal{B}$ until the activation rule is met. When $\alpha(\sigma_t) =$ {\tt true}, all the activity instances in $\mathcal{I}_a^\mathcal{B}$ are executed as a batch, with $\tau_s$ as the current time in $\sigma_t$, and $\tau_c$ determined by the batch type $\beta$. If the process execution ends before activation, any remaining instances are batched.
\end{definition}

Scheduled activation rules can combine \( c^{\text{day}}(\sigma_t) \) and \( c^{\text{hour}}(\sigma_t) \) to trigger the batch at specific day-hour pairs, e.g., matching observed execution patterns:
\begin{equation}\label{eq:combined_scheduled}
c^{\text{sched}}(\sigma_t) \equiv \bigvee_{(d, h) \in \Omega} \Big( c^{\text{day}}(\sigma_t) = d \land c^{\text{hour}}(\sigma_t) = h \Big),
\end{equation}
where \( \Omega \) is the set of day-hour pairs defined by historical data or scheduling needs. For example, batches may start on Mondays at 3:00 PM or Thursdays at 2:00 PM to match high resource availability peaks. Also, activation rules like size-based could be designed considering previous executions of the process:

\begin{equation}\label{eq:batch_size_adjustment}
c^{\text{size}}(\sigma_t) \equiv \lvert \mathcal{I}_a^\mathcal{B} \rvert \geq \theta_v', 
\quad \text{with} \quad 
\theta_v' = \lambda \cdot \frac{1}{|\mathcal{B}_a^{\mathcal{P}}|} \sum_{\mathcal{B} \in \mathcal{B}_a^{\mathcal{P}}} \lvert \mathcal{I}_a^B \rvert,
\end{equation}
where $\theta_v'$ is the allowed batch size threshold, calculated as the average batch size from past executions of activity $a$ in process $\mathcal{P}$, scaled by $\lambda$. Here, {\bf a)} if $\lambda > 1$, the batch size increases; conversely, {\bf b)} if $0 < \lambda < 1$, the size decreases.

Although Definitions 1–3 define batching at the activity level, our optimization approach integrates the policies into a full-process simulation, capturing control-flow semantics, resource hand-offs, and temporal dependencies. Thus, it fully considers the impact of one activity's batching behavior on others, such as delays in enabling or executing downstream activities.

While batching may improve processing efforts and reduce costs, rigid policies that ignore process-specific characteristics can cause inefficiencies. In the following, we explore such scenarios and discuss heuristic interventions.


\subsection{Heuristic Interventions}
\label{sct:heuristics}

Aligned with the primary goal of this paper to optimize costs and times simultaneously, we identified 19 scenarios that can impact one or both dimensions, either directly or indirectly. These scenarios are organized into four categories: waiting time, processing time, cost efficiency, and resource utilization. Each category targets a specific aspect of batch processing that affects cost or time. Then, the proposed heuristic interventions adjust the batching policy to address these scenarios, aiming to enhance the overall process performance.

\begin{heuristic} {\bf -- Waiting Time Related Scenarios}\vspace{+1.0mm}

\item {\sc \textbf{Scenario 1:}} The first instance in a batch typically waits the longest before execution, making it the main contributor to the overall waiting time in the batch. If this first instance waits significantly longer than other instances, e.g., due to low arrival rates or scheduling constraints, the batching activation may be excessively delayed.  {\sc \textbf{Then:}} Add the time-based activation condition $c^{\text{wt-first}}(\sigma_t) \equiv t - \tau_e(i_1) \geq \theta_f'$, where: $$\theta_f' = \lambda \cdot \frac{1}{| \mathcal{B}^{\mathcal{P}}_a |} \sum_{\mathcal{B} \in \mathcal{B}^{\mathcal{P}}_a} \max_{i_k \in \mathcal{I}_a^B} (t - \tau_e(i_k)).$$  This formula adjusts the batch activation threshold $\theta'_f$ by averaging the longest waiting times from previous batches of activity $a$ observed during the process execution. It scales the result by $\lambda$ to control the threshold’s sensitivity to observed delays, e.g., $\lambda = 0.9$ lowers the threshold by 10\%, allowing earlier activation.

\item {\sc \textbf{Scenario 2:}} The last instance in a batch typically waits the least before execution. However, its waiting time can become a bottleneck if the batch is close to reaching the size threshold and the next activity instance is unlikely to arrive soon due to resource unavailability or scheduling constraints. Thus delaying the execution of the entire batch. {\sc \textbf{Then:}} Add the time-based activation condition $c^{\text{wt-last}}(\sigma_t) \equiv t - \tau_e(i_n) \geq \theta'_l$. 
Similar to Scenario 1, this rule targets the last instance instead, adjusting the threshold based on the minimum waiting time: \vspace{-1.5mm} $$\theta_l' = \lambda \cdot \frac{1}{| \mathcal{B}^{\mathcal{P}}_a |} \sum_{\mathcal{B} \in \mathcal{B}^{\mathcal{P}}_a} \min_{i_k \in \mathcal{I}_a^B} (t - \tau_e(i_k)).\vspace{-1.5mm}$$ 

\item {\sc \textbf{Scenario 3:}} When activities are frequently enabled or executed at specific times, but batches are misaligned and triggered at unscheduled intervals outside those timeframes, instances might accumulate, resulting in higher waiting times. {\sc \textbf{Then:}} Add a scheduled activation condition, $c^{\text{sched}}(\sigma_t)$, Equation~\eqref{eq:combined_scheduled}, where \( \Omega \) is the set of high-frequency day-hour pairs identified from historical data. These pairs correspond to execution peaks, scheduling batch activation during periods of high enablement or execution frequencies.

\item {\sc \textbf{Scenario 4:}} If batches are triggered when few resources are available, instances may wait longer for execution due to resource unavailability, thus increasing waiting times. {\sc \textbf{Then:}} Add an activation condition, $c^{\text{sched}}(\sigma_t)$, Equation~\eqref{eq:combined_scheduled}, with a focus on resource availability. Thus, $\Omega$ is the set of day-hour pairs with the highest number of resources available, determined by analyzing the availability patterns of resources observed executing activity instances. Thus reducing the likelihood of delays due to resource unavailability.

\item {\sc \textbf{Scenario 5:}} Large batch sizes require more activity instances to wait before activation, potentially delaying execution and increasing waiting times. {\sc \textbf{Then:}} Reduce the size threshold by Equation~\eqref{eq:batch_size_adjustment} b).
    
\end{heuristic}

\begin{heuristic} {\bf -- Processing Time Related Scenarios}\vspace{+1.0mm}

\item {\sc \textbf{Scenario 6:}} For tasks with high processing times that can be executed in parallel, small batch sizes limit the opportunity for concurrent execution, increasing cumulative processing times, which could be reduced with larger batches. {\sc \textbf{Then:}} Increase batch size threshold, Equation~\eqref{eq:batch_size_adjustment} a), to reduce the cumulative processing time via parallel execution.

\item {\sc \textbf{Scenario 7:}} When tasks in a batch are executed sequentially or with low concurrency, increasing the batch size does not (significantly) reduce cumulative processing times, making batching no more efficient than individual activity execution in terms of processing times. {\sc \textbf{Then:}} Reduce batch size threshold, Equation~\eqref{eq:batch_size_adjustment} b), as batching provides no advantage.

\item {\sc \textbf{Scenario 8:}} If batches are triggered close to periods of resource unavailability, activity instances may experience idle times when the resource becomes unavailable mid-execution, increasing overall processing times. {\sc \textbf{Then:}} Add the scheduled activation condition, $c^{\text{sched}}(\sigma_t)$, Equation~\eqref{eq:combined_scheduled}, where \( \Omega \) is the set of day-hour pairs selected based on one of the following criteria:\vspace{-1.5mm}
        \begin{itemize}
            \item Pairs are chosen to enforce the batch starting at the beginning of a longer resource availability window period, according to their working calendars, to minimize interruptions.
            \item Pairs are selected for the batch to start on timeslots from where the average processing time is less than or equal to the resource availability period, so the batch completes before resource unavailability.
        \end{itemize} 

\item {\sc \textbf{Scenario 9:}} If batch activation occurs without aligning the batch waiting times to resource availability periods, activity instances may start execution during fragmented or short availability windows, causing interruptions and increasing idle times. {\sc \textbf{Then:}} Adjust the time-based activation to align batch execution with optimal processing windows by updating: $c^{\text{wt-first}}(\sigma_t) \equiv t - \tau_e(i_1) \geq \theta'_f$ and $c^{\text{wt-last}}(\sigma_t) \equiv t - \tau_e(i_n) \geq \theta'_l$ where the updated waiting time thresholds are:
    $$
    \theta_f' = \lambda \cdot \frac{1}{|\mathcal{B}_a^{\mathcal{P}}|} \sum_{\mathcal{B} \in \mathcal{B}_a^{\mathcal{P}}} \Delta_w(i_1),
    \quad
    \theta_l' = \lambda \cdot \frac{1}{|\mathcal{B}_a^{\mathcal{P}}|} \sum_{\mathcal{B} \in \mathcal{B}_a^{\mathcal{P}}} \Delta_w(i_n),
    $$
    where:
    $\Delta_w(i_1)$ and $\Delta_w(i_n)$ are calculated by comparing the currently observed waiting times with the start of the most suitable execution window for the batch. This suitability is determined by the resources' availability and the estimated processing time of the batch, aiming to minimize idle times. Then, $\lambda$ is a scaling factor for adjusting the thresholds.
\end{heuristic}

\begin{heuristic}  {\bf -- Cost Related Scenarios} \vspace{+1.0mm}

\item {\sc  \textbf{Scenario 10:}} When batch costs decrease with larger batch sizes, grouping more instances reduces the total cost compared to processing them separately, i.e., analogous to bulk purchasing, where buying in larger quantities reduces the cost per unit. {\sc \textbf{Then:}} Increase batch size threshold, Equation~\eqref{eq:batch_size_adjustment} a), as larger batches can reduce the overall process costs.  

\item {\sc \textbf{Scenario 11:}} High-cost activities might contribute the most to total process costs. Grouping these costly instances into larger batches can amortize shared fixed and variable costs, reducing the average cost per instance and lowering their impact on overall cost. {\sc \textbf{Then:}} Increase batch size threshold, Equation~\eqref{eq:batch_size_adjustment} a), to amortize shared costs of costly activity instances.

\item {\sc \textbf{Scenario 12:}} Frequently executed activities may contribute significantly to total process costs due to repeated fixed and variable costs. Grouping these recurring instances into larger batches may amortize them and minimize their impact on overall process costs. {\sc \textbf{Then:}} Increase batch size threshold, Equation~\eqref{eq:batch_size_adjustment} a), to amortize costs of very frequent activities.

\item {\sc \textbf{Scenario 13:}} Low or mid-cost activities with similar enablement or execution patterns might often be overlooked for batching due to their low contribution to the overall cost. When executed at similar times, grouping these instances into larger batches synchronizes execution, minimizing logistics overhead and thus may improve cost efficiency. {\sc \textbf{Then:}} Increase batch size threshold, Equation~\eqref{eq:batch_size_adjustment} a), for low to mid-cost activity instances with similar execution patterns.

\item {\sc \textbf{Scenario 14:}} Low-cost and low-frequency activities without synchronized patterns might have minimal impact on total process cost. Large batch sizes increase waiting times without saving costs. {\sc \textbf{Then:}} Reduce batch size threshold, Equation~\eqref{eq:batch_size_adjustment} b), as smaller batches do not affect cost efficiency.

\item {\sc \textbf{Scenario 15:}} When increasing batch size does not reduce cost, it may only introduce delays without improving cost efficiency. For example, when larger batches do not reduce fixed and variable costs due to setup costs remaining constant regardless of batch size. {\sc \textbf{Then:}} Reduce batch size threshold, Equation~\eqref{eq:batch_size_adjustment} b), as smaller batches do not affect cost efficiency.
\end{heuristic}

\begin{heuristic} {\bf -- Resource Related Scenarios} \vspace{+1.0mm}

\item {\sc \textbf{Scenario 16:}} High resource utilization above a certain threshold can create bottlenecks, causing delays due to resource unavailability or competing activity assignments. Larger batch sizes may hold resources for extended periods, thus increasing congestion. {\sc \textbf{Then:}} Reduce batch size threshold, Equation~\eqref{eq:batch_size_adjustment} b), to lower resource congestion and minimize waiting times. 

\item {\sc \textbf{Scenario 17:}} Low resource utilization below a certain threshold may indicate underutilization, where resources are idle for long periods, increasing fixed costs relative to unproductive resources. Smaller batch sizes may fail to allocate available resources, reducing cost efficiency. {\sc \textbf{Then:}} Increase batch size threshold, Equation~\eqref{eq:batch_size_adjustment} a), to better utilize available resource capacity. 

\item {\sc \textbf{Scenario 18:}} Activities with high resource allocation variability require frequent switching between resources, increasing setup and transition times. Small batch sizes may add inefficiencies by frequently reallocating resources. {\sc \textbf{Then:}} Increase batch size threshold, Equation~\eqref{eq:batch_size_adjustment} a), to reduce allocation switches and minimize overhead on transition times.

\item {\sc \textbf{Scenario 19:}} Activities with low variability in resource allocation require fewer switches between resources. Large batch sizes may unnecessarily hold resources for extended periods, decreasing overall efficiency. {\sc \textbf{Then:}} Reduce batch size threshold, Equation~\eqref{eq:batch_size_adjustment} b), to release resources sooner, improving allocation flexibility and reducing idle times.
\end{heuristic}


\section{Optimization of Batching Policies}
\label{sec:optimization}

Optimizing batching policies to minimize time and costs in a business process simultaneously requires dynamically adding, updating, or removing activation conditions. However, the search space of activation condition candidates is infinite as time thresholds can take continuous values, and activation rules can combine in numerous ways. Therefore, exhaustively exploring all alternatives is computationally infeasible. The heuristic interventions discussed in Section~\ref{sct:heuristics} address this challenge by reducing the search space to the most relevant scenarios, structuring the search process around relevant combinations, and prioritizing adjustments with potentially the highest impact on cycle time or cost. 

The optimization should also balance the inherent trade-offs between cycle time and cost. Smaller batch sizes reduce waiting times but increase costs, as more instances may add a fixed cost independently. 
Conversely, larger batch sizes reduce costs by amortizing fixed costs across multiple instances but increase times as instances must wait longer to accumulate before triggering the batch. These conflicting objectives prevent finding a single solution that optimizes both dimensions simultaneously. Instead, they require a Pareto front, representing the set of non-dominated solutions where no other option is simultaneously better in cycle time and cost. A solution is on the Pareto front if improving one objective worsens the other. For example, smaller batches achieve a 2-day cycle time for \$500; larger batches reduce costs to \$300 but increase the cycle time to 5 days. Both solutions are optimal as neither is better in both dimensions. This multi-objective optimization problem is formalized by Definition~\ref{def:optimization_problem}.

\begin{definition}
\label{def:optimization_problem}

To minimize the average cycle time and cost, the goal is to find the optimal batching policy for each activity $a \in \mathcal{A}$. This is formulated as a multi-objective optimization problem, where the batching policy for each activity specifies the activation rules and batching characteristics as defined in Definitions~\ref{def:batch} and~\ref{def:batch_activation}. The optimization problem is $\min \frac{1}{D(\mathcal{B})} \left\{ F_1(\mathcal{B}), F_2(\mathcal{B}) \right\}$, with: 
$$
F_1(\mathcal{B}) = \sum_{a \in \mathcal{A}} \sum_{\mathcal{B} \in \mathcal{B}_a^{\mathcal{P}}} \left( \tau_c^\mathcal{B} - \min_{i \in \mathcal{I}_a^\mathcal{B}} \tau_e(i) \right), F_2(\mathcal{B}) = \sum_{a \in \mathcal{A}} \sum_{\mathcal{B} \in \mathcal{B}_a^{\mathcal{P}}} \kappa, D(\mathcal{B}) = \sum_{a \in \mathcal{A}} \sum_{\mathcal{B} \in \mathcal{B}_a^{\mathcal{P}}} |\mathcal{I}_a^\mathcal{B}|
$$
where $F_1(\mathcal{B})$ is the cumulative cycle time, calculated as the sum of cycle times for all batches, from the earliest enablement of any instance in the batch to the completion of the batch, and $F_2(\mathcal{B})$ is the cumulative cost, calculated as the sum of all costs per batch. These values are then normalized by $D(\mathcal{B})$, i.e., the total number of activity instances executed, to minimize the average cycle time and cost per activity. Activity instances executed independently are treated as batches of size 1, triggered as soon as the allocated resource becomes available.
\end{definition}

We follow an iterative data-driven approach to achieve the optimization goals stated in Definition~\ref{def:optimization_problem}. Specifically, we consider event logs representing the execution/simulation of the business process subject to optimization. The approach automatically analyzes these logs to extract sources of inefficiencies, supported by the 19 scenarios described by the heuristic in Section~\ref{sct:heuristics}. For example, waiting time-related scenarios target the activity with high waiting times, ignoring those without delays. Then, they propose actions to reduce it based on different conditions. So, {\sc  \textbf{Scenarios 1-2}} require delays to be batching-related since updating a batching waiting threshold is meaningless on not batched activities. {\sc  \textbf{Scenario 3}} calculates the most frequent activity enablement timeslots to align the execution to these times. Similarly, {\sc  \textbf{Scenario 4}} analyzes the start times of resources assigned to the most delayed tasks, detecting intersections in resource availability to synchronize the execution accordingly. For space reasons, we do not detail how each of the 19 scenarios is automatically discovered and analyzed from the event log. Readers can refer to the source code repository for details.

We rely on simulation to evaluate the impact of each action on the cycle time and cost. If an event log of the process execution exists, the approach discovers a simulation model from it; otherwise, a simulation model must be provided. 
Since batching interventions are evaluated within full-process simulations, the optimization implicitly considers interdependencies among batched activities.


Given a simulation model with certain batching policies attached to its activities, the heuristic interventions in Section~\ref{sct:heuristics} allow us to calculate new (potentially better) sets of batching policies. To explore the space of possible combinations of batching policies, we need to embed these heuristic interventions into meta-heuristics~\cite{Talbi09}. We consider three meta-heuristics. First, we consider Hill-climbing because it is a local search approach suitable for time-constrained scenarios. However, hill-climbing may get stuck in local optima. Accordingly, we also consider simulated annealing, which addresses hill-climbing's limitation while still offering a relatively low computational overhead. Finally, as a representative of a global adaptative search approach, we use reinforcement learning.

\subsection{Hill-Climbing and Simulated Annealing Optimization}
\label{hill_climbing}

The algorithms described in this section are multi-objective variants of hill-climbing and simulated annealing, which are widely used for search optimization problems and are well-documented in the literature~\cite{Talbi09}. These meta-heuristics are adapted to explore the space of batching policies, maintaining a Pareto front of non-dominated solutions, and using Pareto dominance and distance to the front to guide the search across two objectives: cycle time and cost.

Hill-climbing (HC) and simulated annealing (SA) follow the iterative structure described in Algorithm~\ref{algo:unified_optimization}, where each iteration applies heuristic interventions, evaluates the updated batching policies through simulation, refines the Pareto front and select potential candidates, which are perturbated to generate new policies heuristically (lines 2-17). The approach is configurable, allowing the selection of the most problematic scenario, a subset of the most problematic scenarios, or all detected scenarios for intervention (line 10). Since heuristic interventions are independent, multiple scenarios can be executed in parallel to improve computational efficiency. Despite sharing common steps, their search behavior and acceptance criteria differ as described in the following.

\begin{algorithm}[t]
\begin{algorithmic}[1]
\scriptsize
\State \textbf{INPUTS:} {\tt S} $\gets$ {\sc SimulationModel}, {\tt maxSol} $\gets$ {\sc MaxSolutionsCount}, {\tt alg} $\gets$ {\sc SearchStrategy}, {\tt radius} $\gets \epsilon$ (\textcolor{softgreen}{{\sc HillClimbing}}), {\tt temperature} $\gets$ $\infty$, {\tt cF} $\gets$ {\sc CoolingFactor} (\textcolor{red}{{\sc SimulatedAnnealing}})

\State {\tt eLog, cTime, cost} $\gets$ {\sc SimulateBProcess}({\tt S}) \Comment{\textcolor{blue}{Simulate initial process for baseline performance}}
\State {\tt ParetoF} $\gets$  $\{$ {\tt <S, cTime, cost>} $\}$  \Comment{\textcolor{blue}{Initialize Pareto front with the initial solution to optimize}}
\State {\tt Candidates} $\gets$ {\sc InitializeQueue}({\tt alg}) \Comment{\textcolor{blue}{Sorted by distance to Pareto for HC, Random for SA}}

\While{{\tt sCount} $<$ {\tt maxSol} $\land$ {\sc IsNotEmpty}(\tt Candidates)}
    \If{{\tt alg} {\bf is} {\sc \textcolor{softgreen}{HillClimbing}}}
        \State {\tt <{\tt S}, eLog, dist>} $\gets$ {\sc PopBestCandidate}({\tt Candidates}) \Comment{\textcolor{blue}{Retrieve closest candidate}}
    \ElsIf{{\tt alg} {\bf is} {\sc \textcolor{red}{SimulatedAnnealing}}}
        \State {\tt <{\tt S}, eLog, dist>} $\gets$ {\sc PopRandomCandidate}({\tt Candidates}) \Comment{\textcolor{blue}{Retrieve random candidate}}
    \EndIf
    
    \State {\tt Scenarios} $\gets$ {\sc ExtractConflictingScenarios}({\tt eLog}) \Comment{\textcolor{blue}{From heuristics in Section~\ref{sct:heuristics}}}
    
    \For{{\bf each} {\tt sc} in {\tt Scenarios}}
        \State {\tt S'} $\gets$ {\sc ApplyHeuristicAction}({\tt S}, {\tt sc}) \Comment{\textcolor{blue}{Modify batch policy: actions in Section~\ref{sct:heuristics}}}
        \State {\tt eLog', cTime', cost'} $\gets$ {\sc SimulateBProcess}({\tt S'}) \Comment{\textcolor{blue}{Get updated policy performance}}
        \State {\tt dist} $\gets$ {\sc GetEuclideanDistance}({\tt ParetoF}, cTime', cost') \Comment{\textcolor{blue}{Dist to current Pareto}}

        \If{ {\tt dist} $= 0$}  \Comment{\textcolor{blue}{Non-dominated solution, add to Pareto front}}
            \State {\sc UpdateParetoFront}({\tt ParetoF}, {\tt <S', cTime', cost'>}) \Comment{\textcolor{blue}{Update optimal solutions}}
            \State {\sc EnqueueCandidate}({\tt Candidates}, {\tt <S', eLog', 0>}) \Comment{\textcolor{blue}{Consider in future iterations}}
        \ElsIf{{\tt alg} {\bf is} {\sc \textcolor{softgreen}{HillClimbing}} $\land$ {\tt dist} $< {\tt radius}$} \Comment{\textcolor{blue}{Close to Pareto}}
            \State {\sc EnqueueCandidate}({\tt Candidates}, {\tt <S', eLog', dist>}) \Comment{\textcolor{blue}{Save for future iterations}}
        \ElsIf{{\tt alg} {\bf is} {\sc \textcolor{red}{SimulatedAnnealing}} $\land$ {\sc Random}(0,1) $<$ $e^{-\tt dist / temp}$}
            \State {\sc EnqueueCandidate}({\tt Candidates}, {\tt <S', eLog', dist>}) \Comment{\textcolor{blue}{In temperature range}}
        \EndIf
    \EndFor
    
    \If{{\tt alg} {\bf is} {\sc \textcolor{red}{SimulatedAnnealing}}}
        \State {\tt temp} $\gets$ {\tt temp} $\times$ {\tt cF} \Comment{\textcolor{blue}{Reduce temperature to make search more strict}}
        \State {\sc DiscardOverLimit}({\tt Candidates}, {\tt temp}) \Comment{\textcolor{blue}{Remove candidates out of temperature range}}
        \If{ {\tt temp} $< \epsilon$} 
            \State {\tt alg} $\gets$ {\sc \textcolor{softgreen}{HillClimbing}}, {\tt radius} $\gets$ 0 \Comment{\textcolor{blue}{Switch to HC, keep only optimal candidates}}
        \EndIf
    \EndIf
\EndWhile
\State \Return {\tt ParetoF} \Comment{\textcolor{blue}{Return Pareto-optimal solutions}}
\end{algorithmic}
\caption{Unified Hill Climbing and Simulated Annealing Optimization}
\label{algo:unified_optimization}
\end{algorithm}

Hill-climbing (lines 6-7, 15–19) is a local search method that evaluates candidates within a fixed (small) radius from the Pareto front, using Euclidean distance (Algorithm~\ref{algo:unified_optimization}, line 14). This radius mitigates errors from simulation stochasticity, preventing minor variations in the simulation from affecting the search. The algorithm accepts only candidates within this threshold (lines 15–19), incrementally improving solutions for fast convergence. However, restricting exploration to a local neighborhood increases the risk to stop at a local optimum.


In contrast, simulated annealing (lines 8-9, 20–26) is a global search strategy that stochastically accepts candidates based on a temperature-dependent probability distribution. Then, it selects them randomly from the perturbation queue, allowing exploration outside the local neighborhood. It starts with a high temperature, i.e., in the domain of real numbers, meaning it stochastically accepts solutions even if they perform worse, allowing it to explore the search space widely. As the temperature decreases, the chance of accepting worse solutions decreases. Eventually, when the temperature approaches zero, the algorithm behaves like hill-climbing, accepting only optimal candidates. This cooling schedule enables simulated annealing to escape local optima by widening the search space in the early stages while refining solutions in later iterations.
 
\subsection{Reinforcement Learning for Batching Policies}
\label{reinforce_learning}

The reinforcement learning (RL) approach (Algorithm~\ref{algo:rl_optimization}) optimizes batching policies by formulating the problem as a Markov Decision Process (MDP), where the agent iteratively interacts with a simulation environment. States represent the current simulation context, actions correspond to heuristic interventions (Section~\ref{sct:heuristics}), and rewards reflect the effectiveness of these interventions in improving performance. We use Proximal Policy Optimization (PPO), a policy-gradient algorithm that learns a stochastic policy by adjusting its parameters based on observed rewards. PPO also learns a value function that estimates the expected outcome from each state, used to reduce variance and stabilize training. In our setup, the agent applies one intervention at a time and receives a reward after each simulation episode. The reward function evaluates the resulting batching policy based on total cycle time and cost, guiding the agent toward configurations that improve both objectives over time.


\begin{algorithm}[tp]
\begin{algorithmic}[1]
\scriptsize
\State \textbf{INPUTS:}  
\Statex \hspace{1.2em} {\tt S} $\gets$ {\sc SimulationModel}, {\tt mI} $\gets$ {\sc MaxIterations}, 
\Statex \hspace{1.2em} {\tt RLModel} $\gets$ {\sc ReinforcementLearningModel}, {\tt Buffer} $\gets$ {\sc ExperienceBuffer}

\State {\tt eLog, cTime, cost} $\gets$ {\sc SimulateBusinessProcess}({\tt S})  \Comment{\textcolor{blue}{Simulate initial process model}}
\State {\tt ParetoF} $\gets$  $\{$ {\tt <S, cTime, cost>} $\}$ \Comment{\textcolor{blue}{Initialize Pareto front}}  
\State {\tt State} $\gets$ {\sc ExtractState}({\tt eLog, cTime, cost})  \Comment{\textcolor{blue}{Extract initial RL state representation}}

\For{iteration $= 1$ to {\tt mI}}
    \State {\tt Actions} $\gets$ {\sc GetAvailableInterventions}({\tt S})  \Comment{\textcolor{blue}{Retrieve possible heuristic interventions}}
    \If{ {\tt Actions} = $\emptyset$ }  
        \State {\bf Break}  \Comment{\textcolor{blue}{Terminate if no valid actions exist}}
    \EndIf
    \State {\tt ActionMask} $\gets$ {\sc MaskUnavailableActions}({\tt Actions}) \Comment{\textcolor{blue}{Mask invalid interventions}}
    \State {\tt $\alpha$} $\gets$ {\sc SelectAction}({\tt RLModel}, {\tt Actions}, {\tt ActionMask})  \Comment{\textcolor{blue}{Predict best intervention}}
    
    \State {\tt S'} $\gets$ {\sc ApplyHeuristicAction}({\tt S}, {\tt $\alpha$})  \Comment{\textcolor{blue}{Modify batching policy}}
    \State {\tt eLog', cTime', cost'} $\gets$ {\sc SimulateBusinessProcess}({\tt S'})  \Comment{\textcolor{blue}{Evaluate new policy}}
    \State {\tt NextState} $\gets$ {\sc ExtractState}({\tt eLog', cTime', cost'})  \Comment{\textcolor{blue}{Extract new state representation}}

    \State {\tt reward} $\gets$ {\sc EvaluateImprovement}({\tt cTime}, {\tt cost}, {\tt cTime'}, {\tt cost'})  \Comment{\textcolor{blue}{Compute reward}}

    \State {\sc StoreTransition}({\tt State}, {\tt $\alpha$}, {\tt reward}, {\tt NextState}, {\tt Buffer})  \Comment{\textcolor{blue}{Store experience}}

    \If{ {\sc IsParetoOptimal}({\tt S'}, {\tt ParetoF})}   \Comment{\textcolor{blue}{Update Pareto front if non-dominated}}
        \State {\sc UpdateParetoFront}({\tt ParetoF}, {\tt <S', cTime', cost'>})  
    \EndIf

    \If{ {\sc ShouldTrain}({\tt Buffer})} \Comment{\textcolor{blue}{Check training conditions}}
        \State {\sc TrainRLModel({\tt RLModel}, {\tt Buffer})}  \Comment{\textcolor{blue}{Update RL model}}
        \State {\sc ClearExperienceBuffer}({\tt Buffer}) \Comment{\textcolor{blue}{Reset buffer}}
    \EndIf

    \State {\tt State} $\gets$ {\tt NextState}  \Comment{\textcolor{blue}{Update state for next iteration}}

\EndFor
\State \Return {\tt ParetoF}  \Comment{\textcolor{blue}{Return Pareto-optimal solutions}}
\end{algorithmic}
\caption{Reinforcement Learning-Based Optimization}
\label{algo:rl_optimization}
\end{algorithm}

In Algorithm~\ref{algo:rl_optimization}, the RL agent initializes the simulation and extracts a state representation (lines 2–4). It then retrieves feasible heuristic interventions (lines 5–9) and predicts an action (line 10), modifying the batching policy and running a new simulation to assess its impact (lines 11–12). The transition is recorded, and a reward function evaluates effectiveness (lines 13–15). The reward function is defined as $r(S, S') = r_{\text{dom}}$ if $S'$ dominates all solutions in the Pareto front, i.e., it is at least as good in all objectives and strictly better in at least one; $r_{\text{imp}}$ if $S'$ does not dominate but improves at least one Pareto-optimal solution; and $r_{\text{pen}}$ otherwise, where $r_{\text{dom}} > r_{\text{imp}} > r_{\text{pen}}$, so that the agent prioritizes solutions improving the Pareto front while penalizing ineffective modifications. Non-dominated solutions are added to the Pareto front (lines 16–17). The RL model is trained incrementally using past experiences stored in a buffer (line 15). Once enough experience is accumulated, the policy is updated (lines 18–20), allowing continuous adaptation rather than relying on predefined search rules.


Unlike fixed meta-heuristics like hill-climbing or simulated annealing, the RL strategy dynamically balances exploration and exploitation. It explores broadly early in training and refines solutions over time but requires multiple iterations, making it sensitive to poor initialization. Also, training overhead can be significant in large-scale processes. Although RL refines the batching policy via a black-box model, the final output remains a white-box policy (Definitions~\ref{def:batch}-\ref{def:batch_activation}).

\section{Implementation and Evaluation}
\label{sec:evaluation}

To evaluate the quality of the Pareto front approximation, we consider convergence (distance to the optimal Pareto front) and diversity (coverage of the solution space)~\cite{AudetBCLS20}. A good approximation should be close to the optimal front while preserving structural diversity. Thus, we define {\bf EQ1}: How good are the Pareto fronts discovered by our proposal regarding convergence and diversity? Also, to assess the practical impact of our optimization on overall cycle times, we define {\bf EQ2}: How much gain does our approach achieve over the initial process?


We implemented the approaches presented in this paper as a Python prototype. The source code and instructions to run the prototype can be accessed from:~\url{https://github.com/AutomatedProcessImprovement/optimos_v2}.
 
\subsection{Datasets and Experimental Setup}

We relied on 10 real-life business processes with diverse characteristics to answer the evaluation questions. We discovered the simulation models to optimize from their corresponding event logs using the tool {\sc Simod}~\cite{ChapelaCampaLSD25}. 
To evaluate the impact of the heuristic interventions, we used the simulation tool {\sc Prosimos}~\cite{Lopez-PintadoHD22}, supporting the {\sc Simod} models. Table~\ref{tbl:log-description} summarizes the characteristics of these business processes, including the number of resources (RES), activities (ACT), and control-flow gateways, parallel (AND), exclusive (XOR), and inclusive (OR), defining synchronization and decision points. Flow arcs (ARCS) illustrate structural complexity, with some processes having highly unstructured, spaghetti-like behavior. The SIM-Time column reports the average simulation time (in seconds) over ten runs, each executing 1,000 process instances. While we use BPMN for control-flow representation due to the tools we rely upon ({\sc Simod}, {\sc Prosimos}), the proposed method is not BPMN-specific. It applies to simulation models supporting control flow, resource allocation, activity durations, and batch execution.



\begin{table}[tp]
\centering\scriptsize
\caption{Characteristics of the real-life business processes.}\label{tbl:log-description}
\vspace{-1.5mm}
\begin{tabular}{rlcccccccc} \hline
{\bf } & {\bf } & {\bf RES}  & {\bf ACT}  & {\bf AND}  & {\bf XOR}  & {\bf OR}  & {\bf ARCS}  & {\bf SIM-Time}  \\ \hline
1  & ACC  & 561 & 18 & -  & 17  & -  & 47   & 0.62 \\ \hline
2  & BP12  & 58  & 6  & -  & 4   & -  & 14   & 0.3 \\ \hline
3  & BP17  & 148 & 8  & -  & 8   & -  & 23   & 0.4 \\ \hline
4  & BP19  & 311 & 37 & 19 & 67  & 22 & 277  & 16.5 \\ \hline
5  & CALL  & 66  & 8  & -  & 14  & -  & 30   & 0.4 \\ \hline
6  & GOV     & 15  & 98 & 3  & 114 & 2  & 365  & 3.5 \\ \hline
7  & INS     & 125 & 11 & -  & 13  & -  & 33   & 0.8 \\ \hline
8  & PRD    & 48  & 26 & 38 & 70  & -  & 293  & 0.5 \\ \hline
9 & SEP    & 25  & 16 & 4  & 22  & 7  & 82   & 1.5 \\ \hline
10 & TRF   & 29  & 11 & 4  & 12  & 4  & 52   & 1.0 \\ \hline
\end{tabular}
\vspace{-4.5mm}
\end{table}

The event logs used in our evaluation lack batching and cost-related data, i.e., typically unavailable in public records. Accordingly, to avoid assigning arbitrary cost values, we redefine the cost function $\kappa_r$ based on processing times, which could be amortized through parallel execution but increasing waiting times: $\kappa_r = \lambda \cdot \sum_{i \in \mathcal{I}_a^B} p_i$ where $p_i$ represents the processing time of instance $i$, and $\lambda$ is a scaling factor. This formulation excludes (from $p_i$) idle periods due to resource unavailability during activity execution. We excluded processing time from cycle time to avoid redundancy in optimization objectives, considering only waiting and idle times not captured by the cost function. Thus preserving the optimization problem (Definition~\ref{def:optimization_problem}) and ensuring the Pareto front balances execution efficiency and batching effectiveness (productive vs. unproductive time).

To analyze the impact of batching constraints on processing times, we adjust the scaling factor $\lambda$ in the cost function $\kappa_r = \lambda \cdot \sum_{i \in \mathcal{I}_a^B} p_i$ to consider two scenarios. First, \textit{Parallel batching} ($\lambda = \frac{1}{|\mathcal{I}_a^B|}$), where processing time is amortized across instances, so that batch execution time equals the longest individual activity. Second, \textit{Hybrid batching} ($\lambda = \frac{0.5}{|\mathcal{I}_a^B|}$) balances parallel and sequential execution by scaling processing time under sequential execution by a 0.5 factor.

We evaluated these two execution scenarios, {\it parallel} and {\it hybrid}, using the meta-heuristics hill-climbing (HC), simulated annealing (SA), and reinforcement learning (RL). Our heuristic-guided variants, HC+, SA+, and RL+ (Section~\ref{sec:optimization}), incorporate the proposed heuristics, while the baselines, HC-, SA-, and RL-, work without a heuristic guide, relying on the standard search mechanisms of each meta-heuristic via neighborhood random perturbations. Thus quantifying the impact of heuristic-driven optimization against standard meta-heuristic search.


 \subsection{Metrics and Experimental Results}

 Since the actual Pareto front is unknown, we followed \cite{CustodioMVV11} to construct a reference Pareto front ({\tt PRef}), which includes all non-dominated solutions from all algorithm runs. We define {\tt PAprox} as the Pareto front approximated by a single algorithm. To address the experimental questions, we evaluate three metrics:

 \newcommand{\dmin}[2]{\min\limits_{y \in #2} d^2(x, y)}
 \begin{itemize}
     \item {\bf EQ1- Convergence}: The Averaged Hausdorff distance~\cite{AudetBCLS20} measures convergence as the mean root mean squared (RMS) distance between {\tt PAprox} and {\tt PRef}: $\frac{1}{2} \sum_{S \in \{\texttt{PAprox}, \texttt{PRef}\}} \sqrt{\frac{1}{|S|} \sum\limits_{x \in S} \dmin{x}{\bar{S}}}$, where $S$ iterates over both sets and $\bar{S}$ is the opposite set. A lower value means a better convergence.

     \item {\bf EQ1 - Diversity}: Purity~\cite{CustodioMVV11} measures the proportion of {\tt PAprox} solutions included in {\tt PRef}, given by: $|{\tt PAprox} \cap {\tt PRef}| / |{\tt PAprox}|$. A higher purity indicates a better {\tt PAprox}, with a maximum value of 1.

     \item {\bf EQ2 - Gain}: Cycle time difference between the initial solution $S_0$ and {\tt PAprox} is given by: $\mathbb{E}[CT(S_0, n)] - \min \{\mathbb{E}[CT(S, n)] \mid S \in {\tt PAprox} \}$, where $\mathbb{E}[CT(S, n)]$ is the mean cycle time per process case, computed as: $CT(S, n) = \max \{\tau_{c}(e) \mid e \in E_n\} - \min \{\tau_{s}(e) \mid e \in E_n\}$ with $E_n$ as the set of activity instances in the process case $n$, and $\tau_{s}(e), \tau_{c}(e)$ corresponds, respectively, to the start and completion timestamps of an activity instance $e$.

 \end{itemize}
Tables~\ref{tbl:hausdorff-results}-\ref{tbl:cycle-time-results} present the results for convergence, diversity, and gain, highlighting cases where heuristic-driven approaches outperform (or equal) all baselines (green), at least one baseline (blue), underperform all baselines (red), or match some but show no improvement over any baseline (gray).

Regarding {\bf EQ1}, Table~\ref{tbl:hausdorff-results} presents convergence results using the averaged Hausdorff distance. Across parallel and hybrid settings, heuristic-guided approaches outperform at least one baseline in 83\% of cases, confirming their impact on Pareto front convergence. RL+ is the best-performing method overall, surpassing RL- in 80\% of cases and all baselines in 60\%. HC+ follows with a 65\% improvement rate over HC-, while SA+ achieves the lowest gains, surpassing SA- in only 45\% of cases, indicating a weaker contribution of heuristics to SA's search process. The parallel scenario shows better convergence due to more flexibility in amortizing processing times, whereas hybrid execution introduces constraints that reduce the parallelization effects. Finally, columns labeled $++$ and $--$, which compare joint Pareto fronts from all heuristic-guided (++) versus non-guided (--) solutions, show that ++ achieves better convergence in 60\% of cases, confirming that heuristic-driven solutions collectively improve convergence.

\begin{table}[tp]
\centering\scriptsize
\caption{{\bf EQ1 Convergence} -- Averaged Hausdorff Distance Results.}\label{tbl:hausdorff-results}
\vspace{-1.5mm}
\resizebox{\textwidth}{!}{
\begin{tabular}{l|cccccc|cc|cccccc|cc} \hline
{\bf } &
\multicolumn{8}{c|}{\bf Parallel} & 
\multicolumn{8}{c}{\bf Hybrid (factor $\lambda = 0.5$)} \\ \hline
 & 
                {\bf HC+} & {\bf SA+} & {\bf RL+} & {\bf HC-} & {\bf SA-} & {\bf RL-} & {\bf ++} & {\bf --} &
                {\bf HC+} & {\bf SA+} & {\bf RL+} & {\bf HC-} & {\bf SA-} & {\bf RL-} & {\bf ++} & {\bf --}  \\ \hline
ACC    & \cellcolor{blue!25}3.14 & \cellcolor{blue!25}1.97 & \cellcolor{green!25}0.27 & 1.25 & 10.7 & 0.89 & \cellcolor{green!25}0.00 & 0.79 & \cellcolor{blue!25}25.1 & \cellcolor{blue!25}25.2  & \cellcolor{blue!25}25.7 & 25.3 & 0.40 & 26.0 & \cellcolor{red!25}25.1 & 0.37  \\ \hline

BP12    & \cellcolor{green!25}2.74 & \cellcolor{blue!25}4.28 & \cellcolor{green!25}0.09 & 4.37 & 4.95 & 2.78 & \cellcolor{green!25}0.08 & 2.77 & \cellcolor{blue!25}1.79 & \cellcolor{blue!25}1.65  & \cellcolor{blue!25}1.53 & 2.01 & 1.72 & 1.03 & \cellcolor{red!25}1.53 & 1.03 \\ \hline

BP17    & \cellcolor{blue!25}0.27 & \cellcolor{red!25}3.90 & \cellcolor{blue!25}1.62 & 2.93 & 0.69 & 0.04 & \cellcolor{red!25}1.37 & 0.02 & \cellcolor{blue!25}1.94 & \cellcolor{green!25}0.99  & \cellcolor{green!25}0.73 & 1.18 & 5.73 & 2.30 & \cellcolor{green!25}0.12 & 5.88 \\ \hline

BP19    & \cellcolor{blue!25}10.1 & \cellcolor{blue!25}8.46 & \cellcolor{green!25}2.02 & 10.2 & 5.61 & 2.08 & \cellcolor{red!25}1.80 & 0.12  & \cellcolor{blue!25}6.71 & \cellcolor{green!25}5.48 & \cellcolor{green!25}0.00 & 6.30 & 151.6 & 112.7 & \cellcolor{green!25}0.00 & 119.6 \\ \hline

CALL    & \cellcolor{blue!25}8.23 & \cellcolor{blue!25}9.17 & \cellcolor{green!25}0.00 & 6.93 & 15.1 & 5.70 & \cellcolor{green!25}0.00 & 5.70 & \cellcolor{blue!25}60.1 & \cellcolor{blue!25}75.4  & \cellcolor{blue!25}7.44 & 113.6 & 0.60 & 4.87 & \cellcolor{red!25}7.44 & 0.33  \\ \hline

GOV       & \cellcolor{green!25}11.7 & \cellcolor{blue!25}38.8 & \cellcolor{green!25}4.72 & 52.7 & 48.7 & 14.4 & \cellcolor{green!25}2.36 & 14.2  & \cellcolor{green!25}0.29 & \cellcolor{green!25}111.0 & \cellcolor{green!25}53.9 & 126.6 & 132.8 & 111.9 & \cellcolor{green!25}0.17 & 108.41 \\ \hline

INS       & \cellcolor{blue!25}5.80 & \cellcolor{red!25}6.99 & \cellcolor{blue!25}6.19 & 6.43 & 0.30 & 6.20 & \cellcolor{red!25}5.79 & 0.23 & \cellcolor{red!25}2.75 & \cellcolor{red!25}2.48  & \cellcolor{green!25}0.35 & 1.42 & 0.60 & 0.72 & \cellcolor{green!25}0.08 & 0.56 \\ \hline

PRD      & \cellcolor{blue!25}9.05 & \cellcolor{gray!25}12.3 & \cellcolor{green!25}1.09 & 12.3 & 2.55 & 5.53 & \cellcolor{green!25}0.34 & 3.32 & \cellcolor{blue!25}26.9 & \cellcolor{red!25}40.3  & \cellcolor{green!25}3.20 & 39.0 & 39.4 & 16.4 & \cellcolor{green!25}0.00 & 15.9 \\ \hline

SEP    & \cellcolor{blue!25}12.7 & \cellcolor{blue!25}13.0 & \cellcolor{blue!25}5.21 & 16.8 & 5.13 & 16.4 & \cellcolor{green!25}3.66 & 5.25 & \cellcolor{blue!25}57.5 & \cellcolor{blue!25}155.4 & \cellcolor{blue!25}102.7 & 177.8 & 3.11 & 384.8 & \cellcolor{red!25}95.2 & 3.11 \\ \hline

TRF   & \cellcolor{red!25}30.8 & \cellcolor{blue!25}30.3 & \cellcolor{green!25}3.84 & 30.2 & 30.7 & 8.51 & \cellcolor{green!25}3.83 & 8.51 & \cellcolor{red!25}65.8 & \cellcolor{red!25}69.1 & \cellcolor{red!25}63.1 & 31.9 & 26.9 & 13.1 & \cellcolor{red!25}63.1 & 6.08\\ \hline
\end{tabular}
}
\end{table}


\begin{table}[tp]
\centering\scriptsize
\caption{{\bf EQ1 Diversity} -- Purity Results.}\label{tbl:purity-results}
\vspace{-1.5mm}
\resizebox{\textwidth}{!}{
\begin{tabular}{l|cccccc|cc|cccccc|cc} \hline
{\bf } &
\multicolumn{8}{c|}{\bf Parallel} & 
\multicolumn{8}{c}{\bf Hybrid (factor $\lambda = 0.5$)} \\ \hline
 &  
                {\bf HC+} & {\bf SA+} & {\bf RL+} & {\bf HC-} & {\bf SA-} & {\bf RL-} & {\bf ++} & {\bf --} & 
                {\bf HC+} & {\bf SA+} & {\bf RL+} & {\bf HC-} & {\bf SA-} & {\bf RL-} & {\bf ++} & {\bf --} \\ \hline
ACC    & \cellcolor{green!25}0.58 & \cellcolor{green!25}0.06 & \cellcolor{green!25}0.36 & 0.01 & 0.01 & 0.00 & \cellcolor{green!25}0.99 & 0.01 & \cellcolor{green!25}0.66 & \cellcolor{green!25}0.19  & \cellcolor{gray!25}0.00 & 0.03 & 0.13 & 0.00 & \cellcolor{green!25}0.84 & 0.16 \\ \hline

BP12    & \cellcolor{blue!25}0.12 & \cellcolor{blue!25}0.19 & \cellcolor{green!25}0.33 & 0.00 & 0.00 & 0.26 & \cellcolor{green!25}0.72 & 0.28 & \cellcolor{blue!25}0.09 & \cellcolor{blue!25}0.24  & \cellcolor{blue!25}0.22 & 0.00 & 0.00 & 0.46 & \cellcolor{green!25}0.54 & 0.46 \\ \hline

BP17    & \cellcolor{green!25}0.34 & \cellcolor{green!25}0.29 & \cellcolor{blue!25}0.10 & 0.01 & 0.06 & 0.26 & \cellcolor{green!25}0.69 & 0.31 & \cellcolor{green!25}0.26 & \cellcolor{blue!25}0.21  & \cellcolor{green!25}0.26 & 0.00 & 0.00 & 0.26 & \cellcolor{green!25}0.74 & 0.26 \\ \hline

BP19    & \cellcolor{gray!25}0.05 & \cellcolor{red!25}0.00 & \cellcolor{blue!25}0.09 & 0.05 & 0.14 & 0.68 & \cellcolor{red!25}0.14 & 0.86 & \cellcolor{gray!25}0.00 & \cellcolor{gray!25}0.00  & \cellcolor{green!25}1.00 & 0.00 & 0.00 & 0.00 & \cellcolor{green!25}1.00 & 0.00 \\ \hline

CALL    & \cellcolor{gray!25}0.00 & \cellcolor{gray!25}0.00 & \cellcolor{green!25}1.00 & 0.00 & 0.00 & 0.00 & \cellcolor{green!25}1.00 & 0.00 & \cellcolor{gray!25}0.00 & \cellcolor{gray!25}0.00  & \cellcolor{blue!25}0.36 & 0.00 & 0.21 & 0.43 & \cellcolor{red!25}0.36 & 0.64 \\ \hline

GOV       & \cellcolor{gray!25}0.00 & \cellcolor{green!25}0.35 & \cellcolor{green!25}0.29 & 0.00 & 0.06 & 0.29 & \cellcolor{green!25}0.65 & 0.35 & \cellcolor{gray!25}0.00 & \cellcolor{green!25}0.85  & \cellcolor{gray!25}0.00 & 0.00 & 0.08 & 0.08 & \cellcolor{green!25}0.85 & 0.15 \\ \hline

INS       & \cellcolor{green!25}0.64 & \cellcolor{green!25}0.17 & \cellcolor{gray!25}0.10 & 0.10 & 0.17 & 0.10 & \cellcolor{green!25}0.81 & 0.19 & \cellcolor{green!25}0.34 & \cellcolor{green!25}0.26  & \cellcolor{green!25}0.34 & 0.03 & 0.11 & 0.03 & \cellcolor{green!25}0.89 & 0.11 \\ \hline

PRD      & \cellcolor{blue!25}0.26 & \cellcolor{blue!25}0.04 & \cellcolor{blue!25}0.18 & 0.45 & 0.08 & 0.00 & \cellcolor{red!25}0.48 & 0.53 & \cellcolor{green!25}0.61 & \cellcolor{green!25}0.09  & \cellcolor{green!25}0.30 & 0.00 & 0.00 & 0.00 & \cellcolor{green!25}1.00 & 0.00 \\ \hline

SEP    & \cellcolor{green!25}0.33 & \cellcolor{gray!25}0.00 & \cellcolor{green!25}0.59 & 0.04 & 0.04 & 0.00 & \cellcolor{green!25}0.93 & 0.07 & \cellcolor{gray!25}0.00 & \cellcolor{blue!25}0.38  & \cellcolor{gray!25}0.00 & 0.00 & 0.62 & 0.00 & \cellcolor{red!25}0.38 & 0.62 \\ \hline

TRF   & \cellcolor{gray!25}0.00 & \cellcolor{gray!25}0.00 & \cellcolor{blue!25}0.30 & 0.00 & 0.00 & 0.70 & \cellcolor{red!25}0.30 & 0.70 & \cellcolor{gray!25}0.00 & \cellcolor{gray!25}0.00  & \cellcolor{green!25}0.69 & 0.00 & 0.13 & 0.19 & \cellcolor{green!25}0.69 & 0.31 \\ \hline
\end{tabular}
}
\vspace{-4.5mm}
\end{table}

Continuing with {\bf EQ1}, Table~\ref{tbl:purity-results} presents Purity results, assessing the diversity and contribution of each approximated Pareto front to the reference optimal. Here, heuristic-guided approaches outperform at least one baseline in 66\% of cases, showing stronger performance (given more significant differences in the metrics results) than convergence, indicating heuristics help explore a broader range of trade-offs even in cases when they do not find the closest solutions to the optimal front. HC+ achieves the highest diversity improvement individually, surpassing HC- in 50\% of cases, tying in 45\%, and underperforming in 5\%. SA+ and RL+ each outperform their baselines (SA- and RL-) in 50\% and 45\% of cases, tie in 20\%, and underperform in 30\% and 35\%, respectively. Additionally, when considering a joint Pareto front (++ vs. --), diversity improves in 75\% of cases, confirming the impact of heuristics in expanding the solution space.

\begin{table}[tp]
\centering\scriptsize
\caption{{\bf EQ2 Gain} -- Cycle Time Improvement Results.}\label{tbl:cycle-time-results}
\vspace{-1.5mm}
\resizebox{\textwidth}{!}{
\begin{tabular}{l|cccccc|cc|cccccc|cc} \hline
 &
\multicolumn{8}{c|}{\bf Parallel} & 
\multicolumn{8}{c}{\bf Hybrid (factor $\lambda = 0.5$)} \\ \hline
 &   
                {\bf HC+} & {\bf SA+} & {\bf RL+} & {\bf HC-} & {\bf SA-} & {\bf RL-} & {\bf ++} & {\bf --} &
                {\bf HC+} & {\bf SA+} & {\bf RL+} & {\bf HC-} & {\bf SA-} & {\bf RL-} & {\bf ++} & {\bf --} \\ \hline
ACC   & \cellcolor{green!25}0.70 & \cellcolor{green!25}1.26 & \cellcolor{gray!25}0.00 & 0.00 & 0.68 & 0.00 & \cellcolor{green!25}1.26 & 0.68 & \cellcolor{green!25}1.09 & \cellcolor{green!25}1.07 & \cellcolor{gray!25}-0.06 & 1.07 & 0.75 & -0.06 & \cellcolor{green!25}1.29 & 1.07  \\ \hline

BP12    & \cellcolor{gray!25}-0.08 & \cellcolor{green!25}0.17 & \cellcolor{gray!25}-0.13 & -0.13 & -0.13 & -0.13 & \cellcolor{green!25}0.17 & 0.05 & \cellcolor{blue!25}0.08 & \cellcolor{blue!25}0.09  & \cellcolor{gray!25}-0.12 & 0.55 & -0.22 & -0.03 & \cellcolor{red!25}0.08 & 0.55 \\ \hline

BP17    & \cellcolor{gray!25}0.00 & \cellcolor{gray!25}0.00 & \cellcolor{gray!25}0.00 & 0.00 & 0.00 & 0.00 & \cellcolor{red!25}-0.09 & 0.00 & \cellcolor{blue!25}0.41 & \cellcolor{blue!25}0.30 & \cellcolor{blue!25}0.35 & 0.30 & 0.30 & 0.43 & \cellcolor{red!25}0.41 & 0.43 \\ \hline

BP19    & \cellcolor{green!25}30.5 & \cellcolor{blue!25}26.1 & \cellcolor{blue!25}23.9 & 28.8 & 28.4 & 13.9 & \cellcolor{green!25}30.47 & 28.8 & \cellcolor{green!25}7.55 & \cellcolor{green!25}8.78  & \cellcolor{green!25}3.01 & -11.4 & -6.26 & -5.38 & \cellcolor{green!25}3.01 & -6.26 \\ \hline

CALL   & \cellcolor{red!25}672 & \cellcolor{red!25}646 & \cellcolor{green!25}721 & 689 & 689 & 721 & \cellcolor{green!25}721 & 721 & \cellcolor{blue!25}352 & \cellcolor{blue!25}418  & \cellcolor{green!25}728 & 187 & 728 & 726 & \cellcolor{green!25}721 & 721 \\ \hline

 GOV       & \cellcolor{gray!25}5.48 & \cellcolor{green!25}12.8 & \cellcolor{red!25}0.00 & 5.48 & 10.7 & 8.79 & \cellcolor{green!25}12.8 & 10.7 & \cellcolor{green!25}22.7 & \cellcolor{blue!25}21.0  & \cellcolor{green!25}21.1 & 15.3 & 21.1 & 20.5 & \cellcolor{green!25}22.7 & 20.6 \\ \hline

INS       & \cellcolor{gray!25}0.00 & \cellcolor{green!25}0.19 & \cellcolor{gray!25}-0.13 & 0.00 & 0.00 & -0.13 & \cellcolor{green!25}0.19 & -0.13 & \cellcolor{gray!25}0.00 & \cellcolor{gray!25}0.00  & \cellcolor{green!25}0.68 & 0.00 & 0.00 & 0.00 & \cellcolor{red!25}-0.58 & 0.00 \\ \hline

PRD      & \cellcolor{blue!25}1.15 & \cellcolor{blue!25}1.15 & \cellcolor{blue!25}1.15 & 1.15 & 1.78 & 0.00 & \cellcolor{red!25}0.84 & 1.78 & \cellcolor{green!25}1.75 & \cellcolor{green!25}0.55  & \cellcolor{gray!25}-1.19 & 0.55 & 0.26 & -1.19 & \cellcolor{green!25}1.75 & 0.26  \\ \hline

SEP    & \cellcolor{green!25}4968 & \cellcolor{red!25}4863 & \cellcolor{red!25}4913 & 4964 & 4961 & 4915 & \cellcolor{green!25}4968 & 4963 & \cellcolor{green!25}3999 & \cellcolor{red!25}2090 & \cellcolor{blue!25}3318 & 2261 & 3805 & 2414 & \cellcolor{green!25}3999 & 3805 \\ \hline

TRF   & \cellcolor{red!25}760 & \cellcolor{red!25}770  & \cellcolor{blue!25}877 & 772 & 826 & 901 & \cellcolor{red!25}877 & 901 & \cellcolor{red!25}451 & \cellcolor{red!25}400 & \cellcolor{blue!25}760 & 799 & 864 & 738 & \cellcolor{red!25}760 & 864 \\ \hline
\end{tabular}
}
\vspace{-4.5mm}
\end{table}

Finally, to answer {\bf EQ2}, Table~\ref{tbl:cycle-time-results} shows cycle time improvement per process case, measured in hours. While the objective function optimizes individual waiting and idle times vs. processing time per activity, external constraints like multitasking and delays may affect cycle time per case. Despite this indirect impact, not modeled by the objective function, heuristic-guided approaches reduced cycle time in 60\% of cases and kept similar times in 40\% of the remaining cases from the initial process, showing their capacity to improve overall execution cycle times. Note that negative values in the table indicate cycle time increases due to batching. However, values between 0 and -1 are small enough to be attributed to simulation errors, meaning cycle time remains unchanged regarding the original process. Individually, HC+ shows the biggest improvement, beating HC- in 55\% of cases, tying in 25\%, and underperforming in 20\%. RL+ and SA+ each outperform their baselines in 40\% of cases, but RL+ ties in 35\% and underperforms in 25\%, while SA+ ties in 15\% and underperforms in 45\%.


We measured the runtime of each optimization approach across all the 10 datasets. On average, hill-climbing took 1.4 h (min: 0.02 h, max: 2.6 h), simulated annealing 2.7 h (min: 0.02 h, max: 8.6 h), and PPO 11.3 h (min: 6.0 h, max: 23.9 h). Although PPO incurs higher training overhead, all methods are suitable for offline use, where batching policies are computed once and reused. We obtained the results under a configuration allowing up to 10000 solutions per run; reducing this cap would proportionally lower the runtimes. Our experiments include complex processes (e.g., GOV, BP19, with over 100 activities), demonstrating scalability. Despite the high-dimensional search space, local heuristic interventions produce substantial gains with a limited number of iterations.

A more extensive evaluation confirming the findings can be accessed from ~\url{https://github.com/AutomatedProcessImprovement/optimos_v2}.
It includes computational overhead (i.e., total optimization time per method, and time per iteration), number of solutions explored, quality metrics of the obtained Pareto fronts, and other insights.

\section{Conclusion}
\label{sec:conclusion}

We proposed an approach to optimize batching policies. The keystone of the approach is a set of 19 heuristics to balance the trade-offs between waiting time, processing effort, and cost. We integrated these heuristics into three meta-heuristics. The evaluation showed that the proposed heuristics lead to higher quality Pareto fronts relative to a random (non-guided) perturbation approach.

The proposed approach is fully automated and requires minimal technical expertise. Given a simulation model, it autonomously applies domain-specific heuristics and meta-heuristics to explore optimal batching policies, reducing configuration complexity and enabling use in organizations without in-house experts. Future work includes integrating the method into user-friendly interfaces or process mining platforms to support its practical adoption.

Another avenue for future work is to extend the approach to handle batch activation based on data-aware conditions. For example, batch activation could depend on order type, urgency, or production constraints. Yet another avenue is extending the proposed approach to support the optimization of activity prioritization policies and multitasking policies, in addition to batching policies.

\vspace{1.5mm}
\noindent \textbf{Work funded by}: European Research Council (PIX project) and Estonian Ministry of Education \& Research via Estonian Centre of Excellence in AI.

\bibliographystyle{splncs04}
\bibliography{references}

\end{document}